\documentclass{article}

\usepackage{microtype}
\usepackage{graphicx}
\usepackage{subfigure}
\usepackage{booktabs} 

\usepackage{hyperref}
\usepackage{amsmath}
\usepackage{cleveref}
\usepackage{enumitem}
\usepackage{array}
\usepackage{amsfonts}
\usepackage{caption}
\usepackage{comment}
\usepackage{float}
\usepackage{tikz}
\usepackage{amsmath}
\usepackage{etoolbox}
\usepackage{comment}
\usepackage[colorinlistoftodos]{todonotes}
\usetikzlibrary{shapes.geometric}
\usetikzlibrary{shapes.misc}

\DeclareMathOperator*{\argmax}{arg\,max}

\newcommand{\wbnote}[1]{{\color{red}[WB: #1]}}

\newrobustcmd*{\gnpm}{\tikz{\node[regular polygon,
    regular polygon sides=4,
    draw,
    fill=red!90!black,
    minimum size=0.1cm,
    scale=0.5] (0,0) {};}}
\newrobustcmd*{\gnpl}{\tikz{\node[regular polygon,
    regular polygon sides=4,
    draw,
    fill=green!50!black,
    minimum size=0.1cm,
    scale=0.5] (0,0) {};}}
\newrobustcmd*{\gnpk}{\tikz{\node[regular polygon,
    regular polygon sides=4,
    draw,
    fill=blue!80!black,
    minimum size=0.1cm,
    scale=0.5] (0,0) {};}}

\newrobustcmd*{\agnpm}{\tikz{\node[circle,
    draw=black,
    fill=red!90!black,
    inner sep=0pt,
    minimum size=5pt] (0, 0) {};}}
\newrobustcmd*{\agnpl}{\tikz{\node[circle,
    draw=black,
    fill=green!50!black,
    inner sep=0pt,
    minimum size=5pt] (0, 0) {};}}
\newrobustcmd*{\agnpk}{\tikz{\node[circle,
    draw=black,
    fill=blue!80!black,
    inner sep=0pt,
    minimum size=5pt] (0, 0) {};}}

\newrobustcmd*{\cgnpm}{\tikz{\node[isosceles triangle, draw,
    fill=red!90!black,
    minimum size=0.1cm,
    scale=0.4,
    isosceles triangle apex angle=60,
    rotate=90] (0,0) {};}}
\newrobustcmd*{\cgnpl}{\tikz{\node[isosceles triangle, draw,
    fill=green!50!black,
    minimum size=0.1cm,
    scale=0.4,
    isosceles triangle apex angle=60,
    rotate=90] (0,0) {};}}
\newrobustcmd*{\cgnpk}{\tikz{\node[isosceles triangle, draw,
    fill=blue!80!black,
    minimum size=0.1cm,
    scale=0.4,
    isosceles triangle apex angle=60,
    rotate=90] (0,0) {};}}

\newrobustcmd*{\anp}{\tikz{\node[cross out,
    ultra thick,
    draw=black,
    fill=black,
    scale=0.5,
    minimum height=0.1cm,
    rotate=45] {};\node[cross out,
    very thick,
    draw=purple!70!black,
    fill=purple!70!black,
    scale=0.5,
    minimum height=0.1cm,
    rotate=45] {};}}
\newrobustcmd*{\convnp}{\tikz{\node[cross out,
    ultra thick,
    draw=black,
    fill=black,
    scale=0.5,
    minimum height=0.1cm] {};\node[cross out,
    very thick,
    draw=purple!70!black,
    fill=purple!70!black,
    scale=0.6,
    minimum height=0.1cm] {};}}
\newrobustcmd*{\fcgnp}{\tikz{\node[star, draw,
    fill=orange!50!yellow,
    minimum size=0.1cm,
    scale=0.4,
    rotate=0] (0,0) {};}}

\usepackage[accepted]{icml2021}

\icmltitlerunning{Efficient Gaussian Neural Processes for Regression}

\begin{document}

\twocolumn[
\icmltitle{Efficient Gaussian Neural Processes for Regression}



\icmlsetsymbol{equal}{*}

\begin{icmlauthorlist}
\icmlauthor{Stratis Markou}{equal,cam}
\icmlauthor{James R. Requeima}{equal,cam,invenia}
\icmlauthor{Wessel Bruinsma}{cam,invenia}
\icmlauthor{Richard E. Turner}{cam}
\end{icmlauthorlist}

\icmlaffiliation{cam}{Department of Engineering, University of Cambridge, Cambridge, UK}
\icmlaffiliation{invenia}{Invenia Labs, Cambridge, UK}

\icmlcorrespondingauthor{Stratis Markou}{em626@cam.ac.uk}
\icmlcorrespondingauthor{James R. Requeima}{jrr41@cam.ac.uk}

\icmlkeywords{Machine Learning, ICML}

\vskip 0.3in
]



\printAffiliationsAndNotice{\icmlEqualContribution} 

\begin{abstract}
Conditional Neural Processes \citep[CNP;][]{garnelo2018conditional} are an attractive family of meta-learning models which produce well-calibrated predictions, enable fast inference at test time, and are trainable via a simple maximum likelihood procedure.
A limitation of CNPs is their inability to model dependencies in the outputs.
This significantly hurts predictive performance and renders it impossible to draw coherent function samples, which limits the applicability of CNPs in downstream applications and decision making.
Neural Processes \citep[NPs;][]{garnelo2018neural} attempt to alleviate this issue by using latent variables, relying on these to model output dependencies, but introduces difficulties stemming from approximate inference. One recent alternative \citep{bruinsma2021gaussian}, which we refer to as the FullConvGNP, models dependencies in the predictions while still being trainable via exact maximum-likelihood. Unfortunately, the FullConvGNP relies on expensive $2D$-dimensional convolutions, which limit its applicability to only one-dimensional data. In this work, we present an alternative way to model output dependencies which also lends itself maximum likelihood training but, unlike the FullConvGNP, can be scaled to two- and three-dimensional data. The proposed models exhibit good performance in synthetic experiments.
\end{abstract}

\vspace{-0.5cm}
\section{Introduction and Motivation}
\label{intro}

Conditional Neural Processes \citep[CNP;][]{garnelo2018conditional} are a recently proposed class of meta-learning models which promises to combine the modelling flexibility, robustness, and fast inference of neural networks with the calibrated uncertainties of Gaussian processes \citep[GPs;][]{rasmussen2003gaussian}.
CNPs are trained using a simple maximum-likelihood procedure and make predictions with complexity linear in the number of data points.
Recent work has extended CNPs by incorporating attentive mechanisms \cite{kim2019attentive} or accounting for symmetries in the prediction problem \citep{gordon2020convolutional,kawano2021group}, achieving impressive performance on a variety of tasks.

\vspace{-0.20cm}
\begin{figure}[h!]
    \centering
    \begin{subfigure}{}
    \includegraphics[width=0.9\linewidth]{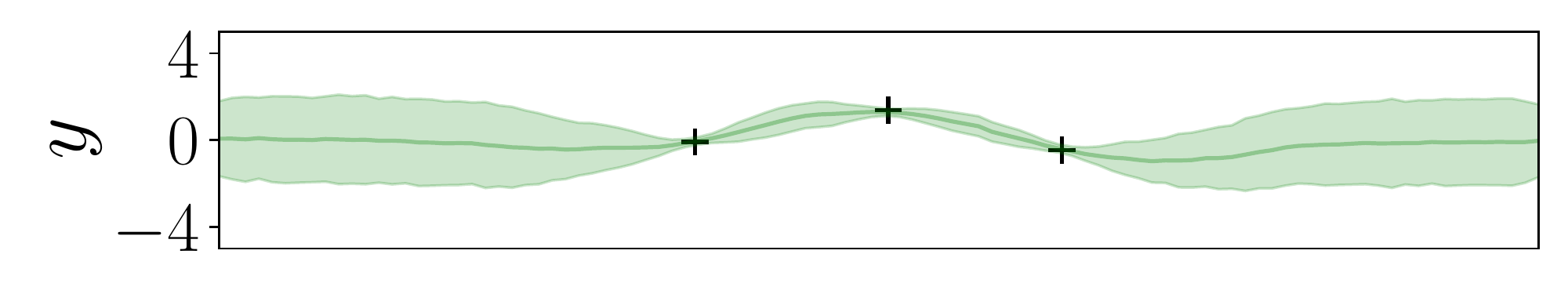}
    \end{subfigure} \vspace{-0.35cm}
    \begin{subfigure}{}
    \includegraphics[width=0.9\linewidth]{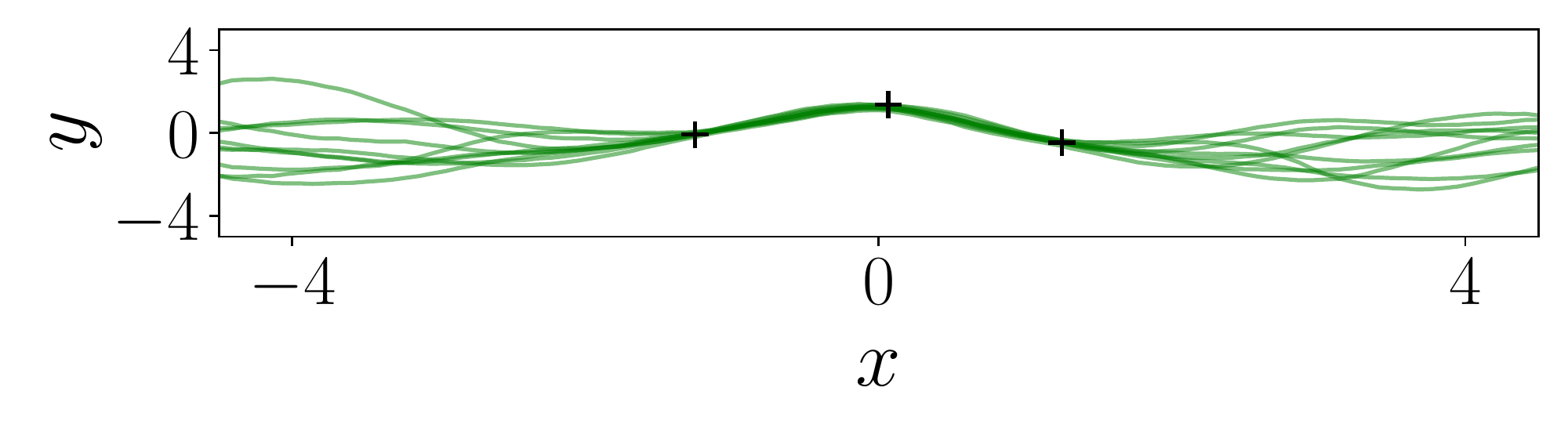}
    \end{subfigure}%
    \vspace{-.5cm}
    \caption{Unlike the ConvCNP (top) which produces only a marginal predictive, the ConvGNP (bottom) provides a correlated predictive, which can be used to draw coherent function samples.\footnote{It's not entirely clear that the top model can only produce marginals. Perhaps a better illustration is by also samping the above model? Then the lack of correlations will be ver clear.}}
    \label{fig:intro}
    \vspace{-0.2cm}
\end{figure}

Despite these favourable qualities, CNPs are limited to predictions which do not model output dependencies, treating different input locations as independent (\cref{fig:intro}).
In this paper, we call such predictions \emph{mean field}.
The inability to model dependencies hurts performance and renders CNPs unable to produce coherent function samples, limiting their applicability in downstream applications.
For example, in precipitation modelling, we might wish to evaluate the probability of the event that the amount of rainfall per day within some region remains above some specified threshold throughout a sustained length of time, which could help assess the likelihood of a flood.
Mean-field predictions, which model every location independently, would assign unreasonably low probabilities to such events. 
If we were able to draw coherent samples from the predictive, however, then the probabilities of these events and numerous other useful quantities could be more reasonably estimated.


To address the inability of CNPs to model dependencies in the predictions,
follow-up work \cite{garnelo2018neural} introduced latent variables, introducing difficulties stemming from approximate inference \citep{le2018empirical,foong2020metalearning}. More recently \citet{bruinsma2021gaussian} introduced a variant of the CNP called the Gaussian Neural Process, hereafter referred to as the FullConvGNP, which directly parametrises the predictive covariance of the outputs.
However, for $D$-dimensional data, the architecture of the FullConvGNP involves $2D$-dimensional convolutions, which are costly, and, for $D > 1$, poorly supported by most Deep Learning libraries.
This work introduces an alternative method to directly parametrise output dependencies, which circumvents the costly convolutions of the FullConvGNP and can be applied to higher dimensional data.

\section{Conditional and Gaussian Neural processes}
\label{gnp}

Following the work of \citet{foong2020metalearning}, we present CNPs from the viewpoint of \textit{prediction maps}. A prediction map $\pi$ is a function which maps (1) a \textit{context set} $(\mathbf{x}_c, \mathbf{y}_c)$ where $\mathbf{x}_c = (x_{c, 1}, \ldots, x_{c, N})$ are the inputs and $\mathbf{y}_c = (y_{c, 1}, \ldots, y_{c, N})$ the outputs and (2) a set of \textit{target inputs} $\mathbf{x}_t = (x_{t, 1}, ..., x_{t, M})$ to a distribution over the corresponding \textit{target outputs} $\mathbf{y}_t = (y_{t, 1}, ..., y_{t, M})$:
\begin{align}\label{eq:pm}
\pi\left(\mathbf{y}_t; \mathbf{x}_c, \mathbf{y}_c, \mathbf{x}_t\right) = p\left(\mathbf{y}_t | \mathbf{r}\right),
\end{align}
where $\mathbf{r} = r(\mathbf{x}_c, \mathbf{y}_c, \mathbf{x}_t)$ is a vector which parameterises the distribution over $\mathbf{y}_t$.
For a fixed context set $(\mathbf{x}_c, \mathbf{y}_c)$, using Kolmogorov's extension theorem \cite{oksendal2013stochastic}, the collection of finite-dimensional distributions (f.d.d.s)
$
    \pi\left(\mathbf{y}_t; \mathbf{x}_c, \mathbf{y}_c, \mathbf{x}_t\right)
$ for all $\mathbf{x}_t \in \mathbb{R}^M,\,M \in \mathbb{N}$ defines a stochastic process
if these f.d.d.s are consistent under (i) permutations of any entries of $(\mathbf{x}_t, \mathbf{y}_t)$ and (ii) marginalisations of any entries of $\mathbf{y}_t$ --- see appendix A. Prediction maps include, but are not limited to, Bayesian posteriors. One familiar example of such a map is the Bayesian GP posterior
\begin{equation} \label{eq:gp}
\pi\left(\mathbf{y}_t; \mathbf{x}_c, \mathbf{y}_c, \mathbf{x}_t\right) = \mathcal{N}\left(\mathbf{y}_t; \mathbf{m}, \mathbf{K}\right),
\end{equation}
where $\mathbf{m} = m(\mathbf{x}_c, \mathbf{y}_c, \mathbf{x}_t)$ and $ \mathbf{K} = k(\mathbf{x}_c, \mathbf{x}_t)$ are given by the usual GP posterior expressions \citep{rasmussen2003gaussian}. Another prediction map is the CNP \cite{garnelo2018conditional}:
\begin{equation} \textstyle \label{eq:cnp}
\pi\left(\mathbf{y}_t; \mathbf{x}_c, \mathbf{y}_c, \mathbf{x}_t\right) = \prod_{m = 1}^M p(y_{t, m} | \mathbf{r}_m),
\end{equation}
where each $p(y_{t, m} | \mathbf{r}_m)$ is an independent Gaussian and $\mathbf{r}_m = r(\mathbf{x}_c, \mathbf{y}_c, x_{t, m})$ is parameterised by a DeepSet\footnote{The DeepSet ensures the posterior map is invariant to permutations of the context set. This is a desirable property in general, which should not be conflated with Kolmogorov consistency.} \cite{zaheer2017deep}. CNPs are permutation and marginalisation consistent and thus correspond to valid stochastic processes.
However, CNPs do not respect the product rule in general --- see appendix A and \citet{foong2020metalearning}. Nevertheless, CNPs and their variants \cite{gordon2020convolutional} have been demonstrated to give competitive performance and robust predictions in a variety of tasks and are a promising class of meta-learning models.

A central problem with the predictive in \cref{eq:cnp} is that is mean field: \cref{eq:cnp} does not model correlations between $y_{t,m}$ and $y_{t,m'}$ for $m \neq m'$. Mean field predictives severely hurt the predictive log-likelihood. 
In addition, one cannot use a mean field predictive to draw coherent function samples.
To remedy these issues, we consider parameterising a correlated multivariate Gaussian
\begin{align} \label{eq:gnp}
\pi\left(\mathbf{y}_t; \mathbf{x}_c, \mathbf{y}_c, \mathbf{x}_t\right) = \mathcal{N}\left(\mathbf{y}_t; \mathbf{m}, \mathbf{K}\right)
\end{align}
where, instead of the expressions for the Bayesian GP posterior, we use neural networks to parameterise the mean $\mathbf{m} = m(\mathbf{x}_c, \mathbf{y}_c, \mathbf{x}_t)$ and covariance $\mathbf{K} = K(\mathbf{x}_c, \mathbf{y}_c, \mathbf{x}_t)$.
We refer to this class of models as Gaussian Neural Processes (GNPs).
The first such model, the FullConvGNP, was introduced by \citet{bruinsma2021gaussian} with promising results. Unfortunately, the FullConvGNP relies on $2D$-dimensional convolutions for parameterising $\mathbf{K}$, which are challenging to scale to higher dimensions. To overcome this difficulty we propose parameterising $\mathbf{m}$ and $\mathbf{K}$ by
\begin{align} \label{eq:gnp1}
\mathbf{m}_i &= f(x_{t, i}, \mathbf{r}), \\ \label{eq:gnp2}
\mathbf{K}_{ij} &= k(g(x_{t, i}, \mathbf{r}), g(x_{t, j}, \mathbf{r}))
\end{align}
where $\mathbf{r} = r(\mathbf{x}_c, \mathbf{y}_c)$, $f$ and $g$ are neural networks with outputs in $\mathbb{R}$ and $\mathbb{R}^{D_g}$ respectively, and $k$ is an appropriately chosen positive-definite function. Note that, since $k$ models a posterior covariance, it cannot be stationary.
\Cref{eq:gnp1,eq:gnp2} define a class of GNPs which, unlike the FullConvGNP, do not require costly convolutions.
GNPs can be readily trained via the log-likelihood
\begin{equation} \textstyle \label{eq:condlik}
\theta^* = \argmax_{\theta} \log \pi\left(\mathbf{y}_t; \mathbf{x}_c, \mathbf{y}_c, \mathbf{x}_t\right),
\end{equation}
also used in \cite{garnelo2018conditional}, where $\theta$ collects all the parameters of the neural networks $f$, $g$, and $r$. In this work, we consider two methods to parameterise $\mathbf{K}$.
The first method is the \texttt{linear} covariance
\begin{equation} \label{eq:innerprod}
\mathbf{K}_{ij} 
= g(x_{t, i}, \mathbf{r})^\top g(x_{t, j}, \mathbf{r})
\end{equation}
which can be interpreted as a linear-in-the-parameters model with $D_g$ basis functions and a unit Gaussian distribution on their weights. This model meta-learns $D_g$ context dependent basis functions, which attempt to best approximate the true distribution of the target given the context. By Mercer's theorem \cite{rasmussen2003gaussian}, up to regularity conditions, every positive-definite function $k$ can be decomposed as
\begin{equation} \textstyle
k(z, z') = \sum_{d=0}^\infty \phi_d(z) \phi_d(z')
\end{equation}
where $(\phi_d)_{d=1}^\infty$ is a set of orthogonal basis functions. We therefore expect \cref{eq:innerprod} to be able to recover arbitrary (sufficiently regular) GP predictives as $D_g$ grows large. Further, the \texttt{linear} covariance has the attractive feature that sampling from it scales linearly with the number of query locations. A drawback is that the finite number of basis functions may limit its expressivity. An alternative method which sidesteps this issue, is to parametrise $\mathbf{K}$ using the \texttt{kvv} covariance:
\begin{equation} \label{eq:kvv}
\mathbf{K}_{ij} = k(g(x_{t, i}, \mathbf{r}), g(x_{t, j}, \mathbf{r})) v(x_{t, i}, \mathbf{r}) v(x_{t, j}, \mathbf{r})
\end{equation}
\begin{figure*}[h!]
    \vspace{-0.1cm}
    \centering
    \includegraphics[width=0.9\linewidth]{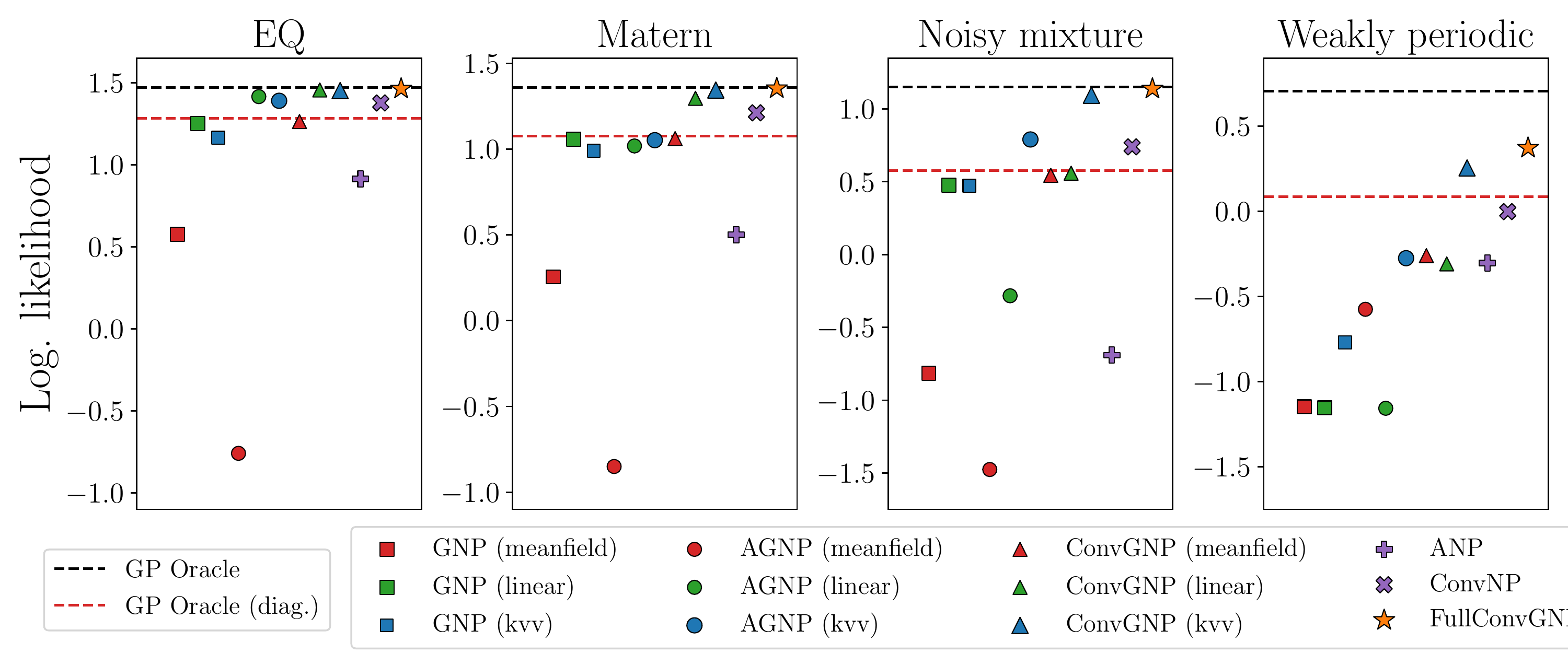}
    \vspace{-0.25cm}
    \caption{Predictive log-likelihood performance of the models, across datasets. The oracle GP performance is shown in dashed black. The dashed red line marks the performance of a modified version of the oracle, in which the off-diagonal terms of the covariance are set to 0.}
    \label{fig:toy-results}
\end{figure*}
\hspace{-0.1cm}where $k$ is the Exponentiated Quadratic (EQ) covariance and $v$ is a neural network with its output in $\mathbb{R}$. The $v$ factors modulate the magnitude of the covariance, which would otherwise not be able to shrink near the context points.
Unlike \texttt{linear}, \texttt{kvv} is not limited by a finite number of basis functions.
%
%
%
%
A drawback of \texttt{kvv} is that the cost of drawing samples from it, scales cubically in the number of query locations, which may impose important practical limitations.

Both \texttt{linear} and \texttt{kvv} leave room for choosing $f$, $g$, and $r$ according to the task at hand, giving rise to a collection of different models of the GNP family. For example, we may choose these to be feedforward DeepSets, giving rise to Gaussian Neural Processes (GNPs); attentive DeepSets, giving rise to Attentive Gaussian Neural Processes (AGNPs); or convolutional architectures, giving rise to Convolutional Gaussian Neural Processes (ConvGNPs). In this work, we explore these three alternatives, proposing the ConvGNP as a scalable alternative to the FullConvGNP. This approach can be extended to multiple outputs, which we will address in future work.

\begin{figure*}[t]
    \vspace{-0.25cm}
    \centering
    \includegraphics[width=1.0\linewidth]{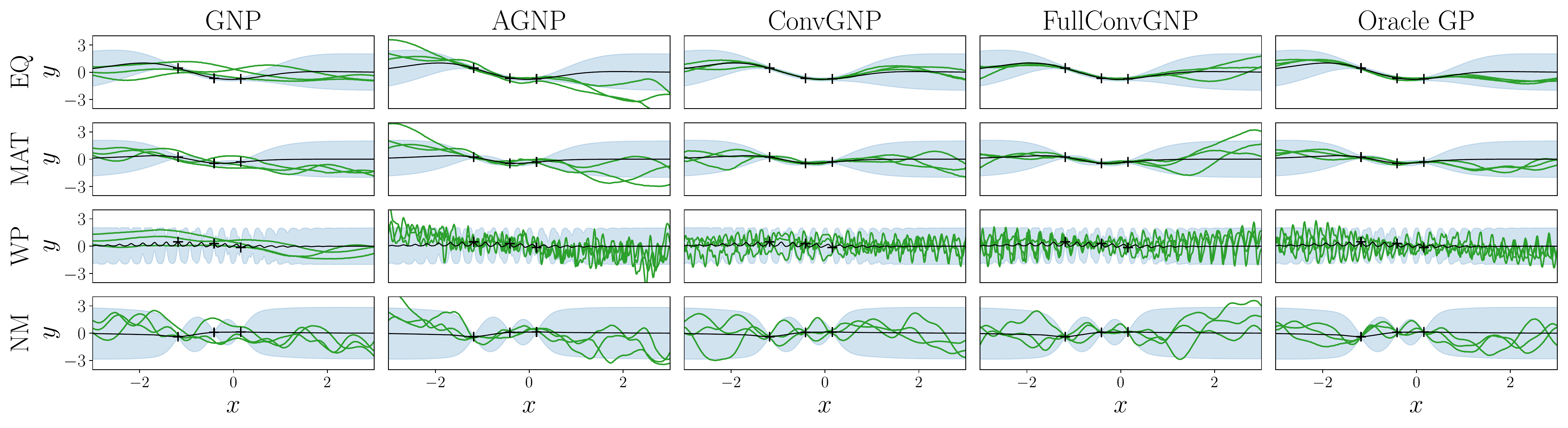}
    \vspace{-0.75cm}
    \caption{Samples drawn from the models' predictive posteriors (green) compared to the ground truth (blue), using the \texttt{kvv} covariance. Note that the models are trained with context and target points uniformly distributed in the $[-2, 2]$ range.}
    \label{fig:post-plots}
    \vspace*{-0.25cm}
\end{figure*}

\section{Experiments}
\label{experiments}

We apply the proposed models to synthetic datasets generated from GPs with various covariance functions and known hyperparameters.
We sub-sample these datasets into context and target sets, and train via the log-likelihood (\cref{eq:condlik}).

We also train the ANP and ConvNP models as discussed in \citet{foong2020metalearning}. These latent variable models place a distribution $q$ over $\mathbf{r}$ and rely on $q$ for modelling output dependencies. Following \citeauthor{foong2020metalearning} we train the ANP and ConvNP via a biased Monte Carlo estimate of the objective 
\begin{equation} \label{eq:latcondlik} \textstyle
\theta^* = \argmax_{\theta} \log \Big[ \mathbb{E}_{\mathbf{r} \sim q(\mathbf{r})}\big[ p\left(\mathbf{y}_t; \mathbf{x}_c, \mathbf{y}_c, \mathbf{x}_t, \mathbf{r} \right) \big]\Big].
\end{equation}
\Cref{fig:toy-results} compares the predictive log-likelihood of the models, evaluated on in-distribution data, from which we observe the following trends.

\textbf{Dependencies improve performance:} We expected that modelling output dependencies would allow the models to achieve better log-likelihoods. Indeed, for a fixed architecture, we see that the correlated GNPs (\gnpl, \gnpk, \agnpl, \agnpk, \cgnpl, \cgnpk) typically outperform their mean-field counterparts (\gnpm, \agnpm, \cgnpm).
This result is encouraging and suggests that GNPs can learn meaningful dependencies in practice, in some cases recovering oracle performance.

\textbf{Comparison with the FullConvGNP:} The correlated ConvGNPs (\cgnpl, \cgnpk) are often competitive with the FullConvGNP (\fcgnp). The $\texttt{kvv}$ ConvGNP (\cgnpk) is the only model, from those examined here, which competes with the FullConvGNP in all tasks. Unlike the latter, however, the former is scalable to $D = 2, 3$ dimensions.

\textbf{Comparison with the ANP and ConvNP:} Correlated GNPs typically outperform the latent-variable ANP (\anp) and ConvNP (\convnp) models, which could be explained by the fact that the GNPs have a Gaussian predictive while ANP and ConvNP do not, and all tasks are Gaussian. Despite experimenting with different architectures, and even allowing for many more parameters in the ANP and ConvNP compared to the AGNP (\agnpl, \agnpk) and ConvGNP (\cgnpl, \cgnpk), we found it difficult to make the latent variable models competitive with the GNPs. We typically found the GNP family easier to train than these latent variable models.

\textbf{\texttt{Kvv} outperformed \texttt{linear}:} We generally observed that the \texttt{kvv} models (\gnpk, \agnpk, \cgnpk) performed as well, and occasionally better than, their \texttt{linear} counterparts (\gnpl, \agnpl, \cgnpl). To test whether the \texttt{linear} models were limited by the number of basis functions $D_g$, we experimented with various settings $D_g \in \{16, 128, 512, 2048\}$. We did not observe a performance improvement for large $D_g$, suggesting that the models are not limited by this factor. This is surprising because, as $D_g \to \infty$ and assuming flexible enough $f$, $g$, and $r$, the \texttt{linear} models should, by Mercer's theorem, be able to recover any (sufficiently regular) GP posterior. From preliminary investigations, we leave open the possibility that the \texttt{linear} models might be more difficult to optimise and thus struggle to compete with \texttt{kvv}. We hope to conduct a more careful study on our training protocol in the future, to determine whether the training method can account for this performance gap, or whether the \texttt{kvv} model is fundamentally more powerful than the \texttt{linear} model.

\begin{figure}[h!]
    \centering
    \includegraphics[width=1.0\linewidth]{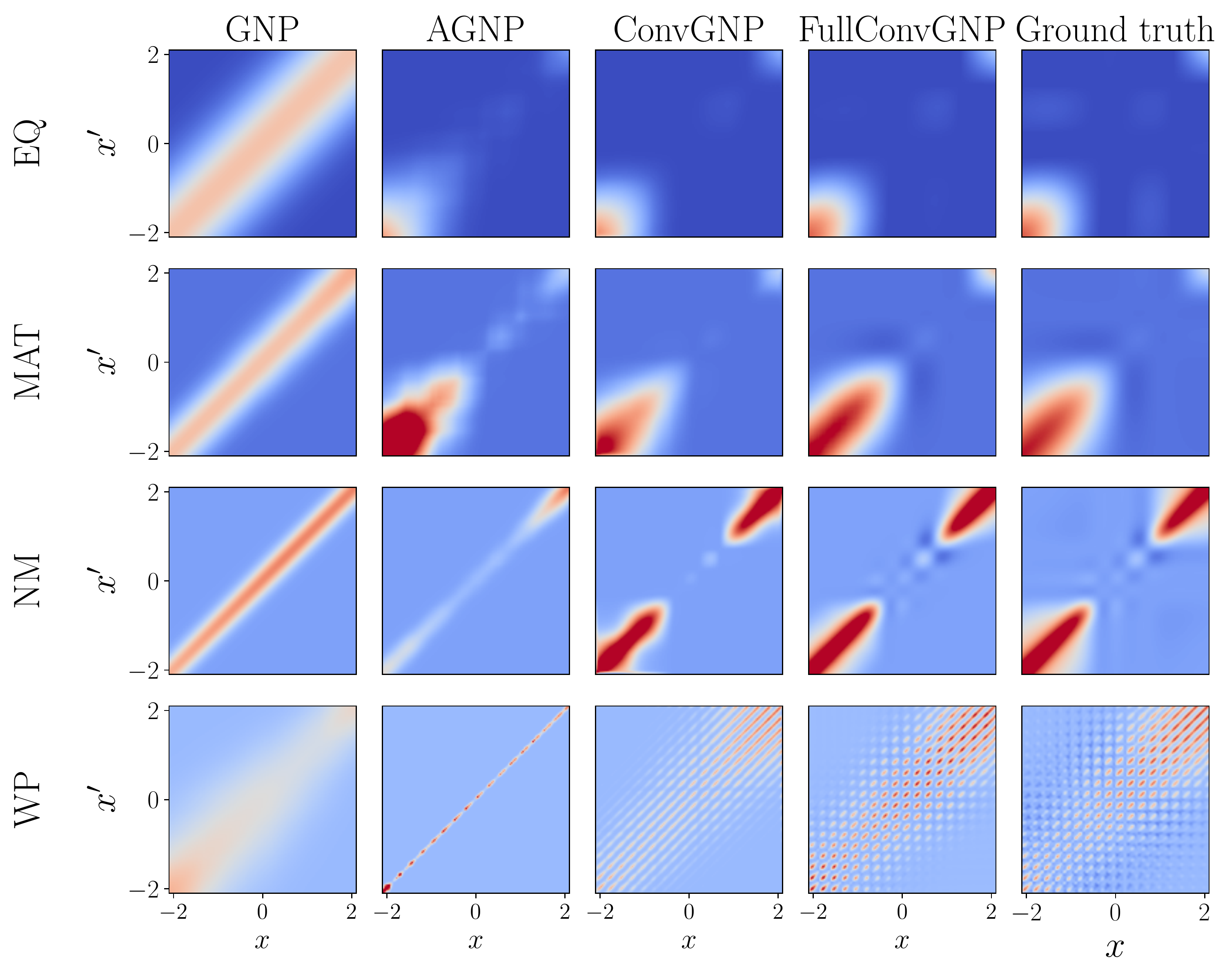}
    \vspace{-0.75cm}
    \caption{Plots of the predictive covariance of the GNP class of models, using the \texttt{kvv} covariance. The FullConvGNP and ground truth covariances are shown for comparison.}
    \label{fig:cov-plots}
    \vspace{-0.5cm}
\end{figure}

\Cref{fig:post-plots} shows samples drawn from the GNP models, from which we qualitatively observe that, like the FullConvGNP, the ConvGNP produces good quality function samples. These samples are consistent with the observed data, whilst maintaining uncertainty and capturing the behaviour of the underlying process. The ConvGNP is the only conditional model (other than the FullConvGNP) which produces high-quality posterior samples. \Cref{fig:cov-plots} shows plots of the models' covariances. Observe that, like the FullConvGNP, the ConvGNP is able to recover intricate covariance structure.

\section{Conclusion and further work}
\label{conclusion}

This work introduced an alternative method for parametrising a correlated Gaussian predictive in CNPs. This approach can be combined with existing CNP architectures such as the feedforward (GNP), attentive (AGNP), or convolutional networks (ConvGNP). The resulting models are computationally cheaper and easier to scale to higher dimensions than the existing FullConvGNP of \citet{bruinsma2021gaussian}, 
whilst still being trainable via exact maximum-likelihood. The ConvGNP outperforms the other conditional and latent-variable models which we consider in this work, with the exception of the FullConvGNP.
%
We found that modelling dependencies in the output improves the predictive log-likelihood over mean-field models. It also allows us to draw coherent function samples, which means that GNPs can be chained with more elaborate downstream estimators. Unlike the ANP and ConvNP models whose predictive is non-analytic, we expect the evaluation of, e.g., Active Learning acquisition functions to be significantly easier and more tractable in GNPs, a use-case we hope to explore in future work.
We also note that, although ConvGNPs exhibit favourable scaling over FullConvGNPs, they still require 2- or 3-dimensional convolutions when applied to higher dimensions, which are also very costly.
We wish to explore ways to reduce this cost, as well as to how other kinds of equivariances such as rotational and reflective equivariance \cite{kawano2021group,holderrieth2020equivariant} can be scaled to higher dimensions, in a computationally cheaper manner. For this, an approach similar to the work of \citet{satorras2021n} in the context of equivariant GNNs is a promising direction. 
We believe that cheap and scalable conditional neural processes for higher dimensional data could be highly valuable in a wide range of applications, including weather and environmental modelling, simulations, graphics and vision.

\clearpage
\section{Acknowledgements}
Richard E. Turner is supported by Google, Amazon, ARM, Improbable and EPSRC grant EP/T005386/1.
\bibliography{references}

\begin{thebibliography}{13}
\providecommand{\natexlab}[1]{#1}
\providecommand{\url}[1]{\texttt{#1}}
\expandafter\ifx\csname urlstyle\endcsname\relax
  \providecommand{\doi}[1]{doi: #1}\else
  \providecommand{\doi}{doi: \begingroup \urlstyle{rm}\Url}\fi

\bibitem[Bruinsma et~al.(2021)Bruinsma, Requeima, Foong, Gordon, and
  Turner]{bruinsma2021gaussian}
Bruinsma, W.~P., Requeima, J., Foong, A. Y.~K., Gordon, J., and Turner, R.~E.
\newblock The gaussian neural process, 2021.

\bibitem[Foong et~al.(2020)Foong, Bruinsma, Gordon, Dubois, Requeima, and
  Turner]{foong2020metalearning}
Foong, A. Y.~K., Bruinsma, W.~P., Gordon, J., Dubois, Y., Requeima, J., and
  Turner, R.~E.
\newblock Meta-learning stationary stochastic process prediction with
  convolutional neural processes, 2020.

\bibitem[Garnelo et~al.(2018{\natexlab{a}})Garnelo, Rosenbaum, Maddison,
  Ramalho, Saxton, Shanahan, Teh, Rezende, and Eslami]{garnelo2018conditional}
Garnelo, M., Rosenbaum, D., Maddison, C.~J., Ramalho, T., Saxton, D., Shanahan,
  M., Teh, Y.~W., Rezende, D.~J., and Eslami, S. M.~A.
\newblock Conditional neural processes.
\newblock \emph{CoRR}, abs/1807.01613, 2018{\natexlab{a}}.

\bibitem[Garnelo et~al.(2018{\natexlab{b}})Garnelo, Schwarz, Rosenbaum, Viola,
  Rezende, Eslami, and Teh]{garnelo2018neural}
Garnelo, M., Schwarz, J., Rosenbaum, D., Viola, F., Rezende, D.~J., Eslami, S.
  M.~A., and Teh, Y.~W.
\newblock Neural processes.
\newblock \emph{CoRR}, abs/1807.01622, 2018{\natexlab{b}}.

\bibitem[Gordon et~al.(2020)Gordon, Bruinsma, Foong, Requeima, Dubois, and
  Turner]{gordon2020convolutional}
Gordon, J., Bruinsma, W.~P., Foong, A. Y.~K., Requeima, J., Dubois, Y., and
  Turner, R.~E.
\newblock Convolutional conditional neural processes, 2020.

\bibitem[Holderrieth et~al.(2020)Holderrieth, Hutchinson, and
  Teh]{holderrieth2020equivariant}
Holderrieth, P., Hutchinson, M., and Teh, Y.~W.
\newblock Equivariant conditional neural processes.
\newblock \emph{CoRR}, abs/2011.12916, 2020.

\bibitem[Kawano et~al.(2021)Kawano, Kumagai, Sannai, Iwasawa, and
  Matsuo]{kawano2021group}
Kawano, M., Kumagai, W., Sannai, A., Iwasawa, Y., and Matsuo, Y.
\newblock Group equivariant conditional neural processes.
\newblock \emph{CoRR}, abs/2102.08759, 2021.

\bibitem[Kim et~al.(2019)Kim, Mnih, Schwarz, Garnelo, Eslami, Rosenbaum,
  Vinyals, and Teh]{kim2019attentive}
Kim, H., Mnih, A., Schwarz, J., Garnelo, M., Eslami, S. M.~A., Rosenbaum, D.,
  Vinyals, O., and Teh, Y.~W.
\newblock Attentive neural processes.
\newblock \emph{CoRR}, abs/1901.05761, 2019.

\bibitem[Le et~al.(2018)Le, Kim, Garnelo, Rosenbaum, Schwarz, and
  Teh]{le2018empirical}
Le, T.~A., Kim, H., Garnelo, M., Rosenbaum, D., Schwarz, J., and Teh, Y.~W.
\newblock Empirical evaluation of neural process objectives.
\newblock In \emph{NeurIPS workshop on Bayesian Deep Learning}, 2018.

\bibitem[Oksendal(2013)]{oksendal2013stochastic}
Oksendal, B.
\newblock \emph{Stochastic differential equations: an introduction with
  applications}.
\newblock Springer Science \& Business Media, 2013.

\bibitem[Rasmussen(2003)]{rasmussen2003gaussian}
Rasmussen, C.~E.
\newblock Gaussian processes in machine learning.
\newblock In \emph{Summer school on machine learning}, pp.\  63--71. Springer,
  2003.

\bibitem[Satorras et~al.(2021)Satorras, Hoogeboom, and Welling]{satorras2021n}
Satorras, V.~G., Hoogeboom, E., and Welling, M.
\newblock E (n) equivariant graph neural networks.
\newblock \emph{arXiv preprint arXiv:2102.09844}, 2021.

\bibitem[Zaheer et~al.(2017)Zaheer, Kottur, Ravanbakhsh, P{\'{o}}czos,
  Salakhutdinov, and Smola]{zaheer2017deep}
Zaheer, M., Kottur, S., Ravanbakhsh, S., P{\'{o}}czos, B., Salakhutdinov, R.,
  and Smola, A.~J.
\newblock Deep sets.
\newblock \emph{CoRR}, abs/1703.06114, 2017.

\end{thebibliography}
\bibliographystyle{icml2021}


\clearpage
\appendix

\section{Consistency}

Here we briefly discuss the consistency of CNPs and GNPs in the Kolmogorov and Bayesian sense. Informally, Kolmogorov's extension theorem (KET) \cite{oksendal2013stochastic} states the following. Suppose that for every list of inputs $\mathbf{x} = (x_{1}, ..., x_{M})$ and corresponding outputs $\mathbf{y}_t = (y_{1}, ..., y_{M})$, 
there is a probability measure $\mu_{\mathbf{x}}$ over $\mathbf{y}_t$.
Suppose that these laws $\{\mu_{\mathbf{x}} : \mathbf{x} \in \mathbb{R}^M, \, M \in \mathbb{R}\}$ are consistent under permutations
\begin{align*}
&\mu_{x_1, \ldots, x_M}(A_{1} \times \ldots \times A_{M}) =\\
&\;=\mu_{x_{\sigma(1)}, \ldots, x_{\sigma(M)}}(A_{\sigma(1)} \times \ldots \times A_{\sigma(M)}),
\end{align*}
and marginalisations
\begin{align*}
&\mu_{x_1,\ldots,  x_K}(A_{1} \times \ldots \times A_{K})\\
&\; = \mu_{x_1,\ldots,  x_M}(A_{1} \times \ldots \times A_{K} \times \mathbb{R} \times \ldots \times \mathbb{R})
\end{align*}
where $1 \leq K \leq M$, $(A_i)_{i=1}^M$ are Borel measurable sets, and $\sigma$ is any permutation, then there exists a stochastic process such that the finite-dimensional distributions of this process coincide with $\{\mu_{\mathbf{x}} : \mathbf{x} \in \mathbb{R}^M, \, M \in \mathbb{R}\}$. We can see directly from their definition in \cref{eq:cnp} that CNPs are both permutation as well as marginalisation consistent. GNPs satisfy permutation consistency because by \cref{eq:gnp,eq:gnp1,eq:gnp2}, a permutation $\sigma$ of the target set indices, i.e. $x_{t, i}, y_{t, i} \to x_{t, \sigma(i)}, y_{t, \sigma(i)}$, leaves the RHS of \cref{eq:gnp} invariant. They are also marginalisation consistent because marginalising out any of the entries of $\mathbf{y}_t$ in \cref{eq:gnp}, gives the same result as querying the GNP at the same set of target points except for the marginalised ones.

However, we stress that consistency of the CNP and GNP in the sense that they satisfy KET is not the same as Bayesian consistency. In particular, the C/GNP predictive posteriors are not expected to satisfy Bayes' rule in general
\begin{align*}
&\pi\left(y^*; (\mathbf{x}_c, x^\dagger), (\mathbf{y}_c, y^\dagger), x^*\right) \pi\left(y^\dagger; \mathbf{x}_c, \mathbf{y}_c, x^\dagger\right) \neq \\ &\pi\left(y^\dagger; (\mathbf{x}_c, x^*), (\mathbf{y}_c, y^*), x^\dagger\right) \pi\left(y^*; \mathbf{x}_c, \mathbf{y}_c, x^*\right),
\end{align*}
where we have used parentheses to denote the inclusion of an additional data to the input/output context sets. Therefore, as \citet{foong2020metalearning} have pointed out, such models do not correspond to a single consistent Bayesian model.

\section{Experimental details}

Each synthetic task consists of a collection of datasets sampled from the same distribution. To generate each of these datasets, we first determine the number of context and target points. We use a random number between $3$ and $50$ of context points and a fixed number of $50$ target points. For each dataset we sample the inputs of both the context and target points, that is $\mathbf{x}_c, \mathbf{x}_t$ uniformly at random in the region $[-2, 2]$. We then sample the corresponding outputs $\mathbf{y}_c, \mathbf{y}_t$ as follows.

\textbf{Exponentiated Quadratic (EQ):} We sample $\mathbf{y}_c, \mathbf{y}_t$ from a GP with an EQ covariance
\begin{align*}
    k_{EQ}(x, x') = \sigma^2_v \exp\left(-\frac{1}{2\ell^2} (x - x')^2\right),
\end{align*}
with parameters $(\sigma_v^2, \ell) = (1.00, 1.00)$.

\textbf{Matern 5/2:} We sample $\mathbf{y}_c, \mathbf{y}_t$ from a GP with a covariance 
\begin{align}
    k_{M}(x, x') = \sigma_v^2 \left(1 + \frac{r}{\ell} + \frac{r^2}{3\ell^2}\right) \exp\left(-\frac{r}{\ell}\right),
\end{align}
where $r = |x - x'|$, with parameters $(\sigma_v^2, \ell) = (1.00, 1.00)$.

\textbf{Noisy mixture:} We sample $\mathbf{y}_c, \mathbf{y}_t$ from a GP which is a sum of two EQ kernels
\begin{align*}
    k_{NM}(x, x') = k_{EQ, 1}(x, x') + k_{EQ, 2}(x, x'),
\end{align*}
with the following parameters $(\sigma_{v, 1}^2, \ell_1) = (1.00, 1.00)$ and $(\sigma_{v, 2}^2, \ell_2) = (1.00, 0.25)$.

\textbf{Weakly periodic:} We sample $\mathbf{y}_c, \mathbf{y}_t$ from a GP which is the product of an EQ and a periodic covariance
\begin{align*}
    k_{WP}(x, x') = k_{EQ}(x, x') \exp\left(-\frac{2\sin^2(\pi |x - x'| / p )}{\ell_p^2}\right),
\end{align*}
with EQ parameters $(\sigma_{v, EQ}^2, \ell_{p}) = (1.00, 1.00)$ and periodic parameters $(p, \ell_{EQ}) = (0.25, 1.00)$.

Lastly, for all tasks we add iid Gaussian noise with zero mean and variance $\sigma_n^2 = 0.05^2$. This noise level was not given to the models, which in every case learned a noise level from the data.

The models were trained for 100 epochs, each consisting of 1024 iterations, at each of which 16 datasets were presented as a minibatch to the models. All models were trained with the Adam optimiser, with a learning rate of $5 \times 10^{-4}$.

\end{document}